# A Cooperative Game-Based Multi-Criteria Weighted Ensemble Approach for Multi-Class Classification**

DongSeong.Yoon, Sogang univ, CS, major: AI

*Abstract*—Since the Fourth Industrial Revolution, AI technology has been widely used in many fields, but there are several limitations that need to be overcome, including overfitting/underfitting, class imbalance, and the limitations of representation (hypothesis space) due to the characteristics of different models. As a method to overcome these problems, ensemble, commonly known as model combining, is being extensively used in the field of machine learning. Among ensemble learning methods, voting ensembles have been studied with various weighting methods, showing performance improvements.

However, the existing methods that reflect the pre-information of classifiers in weights consider only one evaluation criterion, which limits the reflection of various information that should be considered in a model realistically. Therefore, this paper proposes a method of making decisions considering various information through cooperative games in multi-criteria situations. Using this method, various types of information known beforehand in classifiers can be simultaneously considered and reflected, leading to appropriate weight distribution and performance improvement.

The machine learning algorithms were applied to the Open-ML-CC18 dataset and compared with existing ensemble weighting methods. The experimental results showed superior performance compared to other weighting methods.

## I. INTRODUCTION

Recently, artificial intelligence (AI) has been making significant strides in various fields, backed by advancements in diverse methodologies, hardware development, interdisciplinary research, and trials across different domains[1]-[5]. With these technological advances and applications, the challenges AI aims to address are becoming increasingly complex, including issues like data imbalance, high-dimensional noise, and overfitting. Among the methods employed to address these challenges, ensemble learning, commonly referred to as model combining, has been widely adopted[1]-[5].

Ensemble learning has been an integral part of modern technological approaches and features prominently in machine learning research. It is recognized as one of the four primary directions in machine learning research[6][7], frequently mentioned as one of the four approaches to handling imbalanced data[8], and stands alongside deep learning as a dominant field in machine learning[9].

Within the latest techniques for handling complex data, ensemble methods are particularly noteworthy. Key techniques include multimodal learning, which involves training on data from multiple channels, and clustering, where data is categorized into clusters. A crucial element in these techniques is the combination of classifiers, or ensemble[10][11].

The definition of multimodal varies depending on the perspective. It can refer to the combination of data streams such as video and audio, or, from a statistical standpoint, it can mean having multiple modes (peaks)[12][13]. This translates to a combination of features that are statistically significant, not much different from the traditional method of combining relevant features to enhance classification accuracy. In other words, from the perspective of features, it involves combining variously processed data values in different ways. Multimodal techniques can be divided into two types: one that normalizes specific features and inputs them into the same model, and another that combines already made decisions into a new model[14][15]. The latter can be seen as a form of ensemble learning from a decision-making perspective[16][17].

Clustering involves categorizing data into groups based on features. The popular K-means algorithm, for instance, is prone to random variability in its iterations. To overcome this instability, the algorithm is run multiple times, and the most frequent outcome is chosen, which essentially makes the final decision a form of vote-based ensemble[18]. The rationale behind the pervasive use of ensembles is their ability to improve the performance of individual classifiers and their algorithm-independent nature[19]. The Condorcet jury theorem, which posits that under certain preconditions, combining individual estimates enhances outcomes, supports the use of ensembles[20].

Considering ensemble systems, mathematical foundations exist, but the reflection of real-world complexities is arguably one of the primary reasons for their use. In real-world scenarios, especially in complex or significant matters, ensembles are almost always utilized. Examples include consulting experts, convening committees, and conducting votes[21]. Many ideas in machine learning originate from mimicking real-world practices.

In summary, considering ensemble systems involves a fusion of information processing and knowledge combination. Here, the concept of combination is essentially an ensemble. Therefore, ensemble learning is regarded as a perspective in explaining machine learning. Machine learning involves models acquiring knowledge to extract desired information from data, where data and knowledge interact. Preprocessing data for model training, known as feature engineering, presupposes statistical techniques or model-generated features. In this context, ensembles use information processed by knowledge as data, distinguishing them as a type of feature engineering. The distinction between feature engineering and ensemble methods lies in the perspective; many studies in machine learning ultimately revolve around how to process information and combine knowledge.

There are various methods of ensemble learning, each with its own advantages and disadvantages. To introduce traditional ensemble methods, they can be broadly classified into homogeneous and heterogeneous ensembles.

A typical example of a homogeneous ensemble includes bagging[23]-[26], where different classifiers of the same type are trained using randomly extracted datasets, and boosting[27],[28], which involves assigning weights to the dataset. Usually, various modifications of these algorithms are used. Furthermore, in the extension of these methods, dropout, a technique used in neural networks, randomly deactivates nodes during training to create various classifiers and uses all nodes during final prediction, which can be considered an ensemble method[29],[30]. As the name "homogeneous ensemble" implies, these methods combine classifiers of the same algorithm, resulting in a limitation of diversity benefits due to the ensemble not being formed through various methodologies[31].

Heterogeneous ensembles, which this paper focuses on, commonly use stacking (Stack Generalization) and voting as their methods. Stacking, which utilizes a meta-classifier, is a technique where each classifier's output is used as a feature for the meta-classifier to learn. Although generally considered to offer the best performance, it requires separate training and tends to be slower due to increased computational load, with overfitting as a potential downside[32],[33].

The voting method, a relatively intuitive and simple committee approach, combines the results of each classifier according to voting rules. There are various voting rules, and the final result of bagging, for instance, is determined by such a vote. The rationale behind using voting is that it can achieve performance unattainable by a single classifier[34]. This voting method has led to the proposal of numerous methods based on different perspectives of voting rules and weighting methods.

In the case of weighted voting ensembles, the method is straightforward, easy to understand, and effective, enhancing the robustness and accuracy of classifiers[35]-[37]. Therefore, research on performance improvement through voting methods has been actively ongoing, leading to the development of various combination methods. These methods are primarily based on model performance metrics like error rate and accuracy, with most methods relying on a single criterion.

However, no single performance metric can fully represent a model's characteristics. Commonly used evaluation metrics include statistical concepts like accuracy (ACCURACY), positive predictive value (PPV), sensitivity, specificity, and negative predictive value (NPV). In an ideal classifier, all these metrics would yield a value of 1, but such a perfect classifier does not exist in real-world problems, and trade-offs between these metrics are often present.

Moreover, focusing on only one of these evaluation metrics can be inadequate, as depending on the data distribution, some metrics may be meaningless while others are significant. This issue is particularly prominent in class-imbalanced data, and depending on the problem definition, the environment also needs to be considered, such as when sensitivity is prioritized due to differences in error costs.

In multi-class classification, the considerations increase compared to binary classification, as the evaluation methods are viewed differently for each class, leading to varying measurements even for the same evaluation metric from the perspective of each class.

There are not many studies that reflect these various considerations. Existing weighted ensemble methods typically average class-specific evaluation metrics and combine them, use the sum of the diagonal of the confusion matrix, or count errors. These methods may not sufficiently reflect the information of the models forming the ensemble. Therefore, a method is needed that can consider various available information while maintaining the characteristics to adjust the importance of information and calculate weights accordingly.

As the existing methods relied on a single evaluation metric, this paper aims to apply a method that considers multiple pieces of information obtainable from models while adjusting their importance. In this process, cooperative game theory is used, which offers two advantages.

Firstly, it can reflect the complexity of the real world. Weighted voting ensembles, akin to committee-based decision systems, require multifaceted consideration of factors that can influence decisions, such as members' influence and expertise. Real-world committees also devise various procedures and rules to reflect this complexity, but traditional weight distribution methods use only one metric.

The proposed method, compared to traditional ones, reflects the complexity of real-world problems more diversely. Considering multi-criteria situations allows for evaluating each classifier's features from various angles, easily adding different evaluation metrics to weight consideration, and reflecting classifier characteristics that traditional methods fail to represent. This aligns with the core objective of ensembles: to reflect diversity[6].

Secondly, game theory offers intuitive and easily understandable interpretations. That is, it is easy to understand and explain how information is evaluated and weighted, and necessary adjustments to the weights of information can be made easily for modeling.

The issue of group decision-making considering multiple criteria has been continuously researched in the fields of economics and game theory. More specifically, these problems, when treating weights as resources, can be viewed as resource allocation problems in group decision-making and multi-criteria situations. Various specialized analysis methods and solutions exist for these problems, researched extensively in economics and game theory.

Therefore, applying cooperative games to ensemble problems is the most suitable solution for the issue. When applied to the problem, using cooperative games in ensembles corresponds to a situation where classifiers come together for group decision-making, aiming to increase the joint benefit.

## II. RELATED RESEARCH

Research on performance improvement through voting-based ensembles has been actively ongoing, and several classifications of this approach are possible based on different perspectives[38]-[41]. Numerous approaches have been explored, including Bayesian methods[42],[43], decision templates[44], local accuracy-based dynamic ensembles[45], information combination[46],[47], expert networks[42],[48], majority through aggregation[49], evidence-based approaches[50],[51], and more. To briefly summarize these voting ensembles, the notation is as follows:

In a situation with $n$ classifiers and $m$ classes, $E = \{C_1, C_2, \ldots, C_n\}$ represents the ensemble of classifier $C_i$, $a_i$ denotes the accuracy of classifier $i$, $e_i^j$ is the error rate of classifier $i$ for class $j$, and $o_i^j$ represents the output of classifier $i$ for class $j$.

According to the literature, the outputs of classifiers are divided into three types: hard (Crisp), soft (Fuzzy), and possibilistic. The output appears as a vector $[\alpha_1, \ldots, \alpha_n]$, and the differences in each type of output are as follows: For hard (Crisp) output, commonly known as classification results, it is denoted deterministically as $[0, \ldots, 1, \ldots, 0]$ and can be represented as $Ho_i^j \in \{0,1\}, j \in 1,2,\ldots,m$. Soft (Fuzzy) output includes a probabilistic interpretation of output, represented as probabilities of belonging to each class, such as $[0, 0.2, \ldots, 0.5, \ldots]$. It is defined as $So_i^j \in [0,1], \sum_{j=1}^{m} So_i^j = 1$. Lastly, Possibilistic output represents the fitness of a sample for a class and can be denoted as $Po_i^j \in [0,1], \sum_{j=1}^{m} Po_i^j > 0$[44],[52].

In weighted ensemble methods, hard and soft outputs are primarily used, with the difference in outputs leading to simple majority and soft voting methods. Simple Majority Voting (SMV) ensemble, which uses hard output, can be represented as follows[39]:

$$E_{SMV} = \underset{j=1,\ldots,m}{\mathrm{argmax}} \sum_{i=1}^{n} r_i Ho_i^j, r_i = \frac{1}{n} \quad (1)$$

Simple Average Voting (SAV) uses soft output for voting and can be represented as[39]:

$$E_{sav} = \underset{j=1,\ldots,m}{\mathrm{argmax}} \sum_{i=1}^{n} r_i So_i^j, r_i = \frac{1}{n} \quad (2)$$

In addition to these, there are various weighted and unweighted voting methods based on different voting rules,

with the majority of weighted voting ensembles being based on these two types[40],[41]. Here, $r_i$ represents the weight, and the method of determining the weight leads to different weighted voting ensemble approaches. The existing weighted distribution methods selected for comparison in this paper are SWV, RSWV, BWWV, QBWWV, WMV, briefly introduced as follows[38]-[41]:

Simple Weighted Vote (SWV) bases its weights on the predictive performance of classifiers, defined as[40],[41]:

$$r_i = \frac{a_i}{\sum_{i=1}^{n} a_i} \quad (3)$$

WMV (logodds) uses logarithms to determine weights, defined as[38],[40]:

$$r_i = \log\left(\frac{a_i}{1 - a_i}\right) \quad (4)$$

Re-Scaled Weighted Vote (RSWV) assigns a weight of 0 to classifiers with performance below n/m, thus excluding low-performing classifiers from the ensemble, with weights determined as follows[40],[41]:

$$a_i = \max\left\{0, 1 - \frac{m \cdot e_i}{n(m-1)}\right\} \quad (5)$$

Best-Worst Weighted Vote (BWWV) is defined as[40],[41]:

$$a_i = 1 - \frac{e_i - e_B}{e_W - e_B} \quad (6)$$

Quadratic Best-Worst Weighted Vote (QBWWV) is defined as[40],[41]:

$$a_i = \left(\frac{e_W - e_k}{e_W - e_B}\right)^2 \quad (7)$$

These weighting methods depend on a specific performance metric, which may not adequately represent the characteristics of classifiers in an ensemble. Therefore, this research aims to solve this issue by considering multi-criteria situations and applying game theory to evaluate classifiers and distribute weights.

Numerous studies have attempted to apply game-theoretic concepts to ensemble and machine learning. For example, in ensemble clustering, game-theoretic approaches have been used to reduce time and space complexity[53], and in image classification, a weighted majority rule (WMR) based on cooperative games has been employed to select synergistic classifiers for feature selection, classification, and decision fusion in ensembles[54]. Attempts have also been made to improve outcomes by designing voting rules and adjusting weights based on game theory in ensemble decision-making[38].

The concept of pruning ensembles using game theory has also been explored. Evolutionary game theory has been used for pruning to select classifiers[55], and in activity recognition, cooperative game theory has been employed to reduce the number of features from various sensors, using the influence of features to minimize them and inputting the classification results from boosted decision trees into SVMs for reduced computational load[56]. There have also been attempts to use game theory for phoneme selection in speech recognition[57] and to weight specific features[58].

Thus, many studies utilize game theory because it excels in analyzing social phenomena, which are the inspiration for ensembles. Game theory is the most suitable for analyzing and designing given environments, making it an ideal choice for addressing issues in ensemble learning.

## III. GAME THEORY AND MCDM

Game theory analyzes game situations where multiple agents come together to make decisions and receive predetermined rewards based on the outcomes. Many real-world problems correspond to this theory.

Game theory is broadly divided into two types: cooperative games and non-cooperative games. Cooperative games analyze situations where players can form coalitions and enter into binding agreements voluntarily, aiming to share limited resources efficiently and fairly. Non-cooperative games, on the other hand, analyze outcomes based on mutual influence in situations without binding constraints beyond the game's rules, and aim to derive rational decision-making. The weighted ensemble used in this paper falls under cooperative games, as it considers the performance of classifiers to distribute weights that can enhance overall performance. Therefore, the design of weight distribution is based on cooperative game theory.

Previous studies have often selected one performance metric of the base classifiers influencing the ensemble's performance. However, a pre-trained classifier can provide various types of information, and the importance of specific information can vary depending on the situation. There is a need for a method that can integrate and reflect these types of information in the weights by assigning importance to specific information.

Specifically, a method is needed that can scale various types of information appropriately according to the situation, allowing for comparison while simultaneously considering them, followed by a comprehensive evaluation and assigning credibility based on it. An appropriate method for such situations, extensively researched in the field of economics based on game theory, is the VIKOR method. This method involves generating a global evaluation by considering all criteria.

### A. MCDM-VIKOR

#### Number

MCDM (Multi-Criteria Decision Making), widely used in economics, shares roots with game theory as a solution to problems requiring consideration of multiple factors, i.e., in multi-criteria situations[59]. Therefore, it has considerable similarities with negotiation solutions, representative of game solutions[60]. Consequently, researches often combine and compare these approaches depending on the suitability of the problem[60]–[70], and although rare, there are attempts to integrate these into multi-criteria negotiation solutions[71]. In the context of applying cooperative games to machine learning, depending on the perspective, either negotiation solutions or MCDM may be more suitable[72]. This is why this paper seeks to apply MCDM based on cooperative games.

In this paper, we aim to use the VIKOR (VIseKriterijumska Optimizacija I Kompromisno Resenje) method, a cooperative-based approach. VIKOR was developed to solve situations where the ideal solution must be chosen while considering several conflicting and differently unit-based criteria. Specifically, it is a point solution method defining global criteria by calculating distances from an ideal point[66]-[74]. VIKOR method cases based on game theory are readily found in the literature[75],[76].

Typically, in problems where a decision must be made among alternatives, there is more than one factor to consider. Even when purchasing an item, one must consider the seller's reliability, price, waiting time, etc. Although it would be ideal to satisfy all the best conditions, such ideal cases are rare in decisions involving multiple criteria. MCDM was developed to solve such problems, used to prioritize alternatives when multiple factors, like feature selection or material choice in industrial settings, must be considered[77],[78].

Here, we consider assigning weights that consider each performance indicator and assigning weights considering performance indicators for each class. When deciding on an alternative with multiple evaluation criteria, let's say the alternatives are n and the criteria are m. Moreover, let the best alternative for criterion m be $a\_m^*$, and the worst be $a\_m^-$; here, $a\_m^n$ is the value of alternative n for criterion m.

The method to calculate $S\_n$ in the VIKOR method is as follows, representing a comprehensive evaluation across all criteria and the expected opportunity loss[66]:

$$S_n = \sum_{j=1}^{m} \frac{p_j (a_j^* - a_j^n)}{a_j^* - a_j^-} \qquad (8)$$

The method to calculate $R\_n$ is as follows, representing the biggest flaw, or the maximum risk of an alternative[66]:

$$R_n = \max_j \left[ \frac{p_j (a_j^* - a_j^n)}{a_j^* - a_j^-} \right] \qquad (9)$$

$Q\_n$ values are determined considering these $S\_n$ and $R\_n$ values. $p\_j$ represents the weight for each criterion, and the $Q\_n$ value is determined by the weight $v$, which indicates the focus between S and R values. Typically, a value of 0.5 is preferred, with $(v < 0.5)$ showing unstable performance changes. The formula for determining $Q\_n$ is as follows[66],[79]:

$$Q_n = \frac{v(S_n - S^*)}{S^- - S^*} + (1 - v)\frac{(R_n - R^*)}{(R^- - R^*)} \qquad (10)$$

The VIKOR method involves deciding rankings using these $Q\_n$ values.

### B. Solutions in Cooperative Games: Values
#### B-1. Axioms of Values in Cooperative Game Theory

Values in cooperative game theory are characterized by several axioms. Some of the key axioms are:

Efficiency (E): The total of all values must equal the entire amount of the resource. This means there shouldn't be any leftover resources when dividing a resource among players. It

is defined as[80]:

$$\sum_{i \in S} \phi_i(u) = v(S),$$

$$\text{s. t. }, \forall S \in N, u \in v(S), i \in S \qquad (11)$$

Additivity (AD): In a function, the sum of the values should yield the same result. It is expressed as[81]:

$$\phi(j + k) = \phi(j) + \phi(k),$$

$$\text{s. t., } \forall S \in N, \ j, k \in v(S) \qquad (12)$$

Linearity (L): An extension of additivity, where scalar multiplication is added to the condition. It is defined as[80]:

$$\phi(j + k) = \phi(j) + \phi(k) \text{ and } \phi(\alpha \cdot k)$$
$$= \alpha \cdot \phi(k),$$
$$\text{s. t. } \forall S \in N, \ j, k \in v(S), \qquad \alpha \in \mathbb{R} \qquad (13)$$

Symmetry (S): Players with the same contributions should receive equal evaluations. Also known as Equal Treatment Property (ETP), it's defined when the following condition is met[80]:

$$\phi_a(u) = \phi_b(u),$$
$$\text{s. t. }, \forall S \in N, \qquad u \in v(N), \qquad \forall a, b \in N \qquad (14)$$

Anonymity (A): Players with the same contributions should receive equal evaluations regardless of order. It is defined when the following condition is met[80]:

$$\phi_a(u) = \phi_{\pi(a)}(\pi u),$$
$$\text{s. t. }, \text{permutation } \pi: N \to N, \qquad u \in v(N),$$
$$\forall a \in N \qquad (15)$$

The philosophy behind value distribution can broadly be summarized into two categories: prioritizing individual value and emphasizing group value. For example, the Shapley value focuses on a player's marginal contributions, emphasizing individual value, while equal division (one-nth for all) considers the group's value without individual contributions[82].

Most values attempt to find a solution between these two concepts, excluding the individual-focused Shapley value and the group-focused egalitarianism. One aspect to consider is the axiom of Null Player Out (NPO) for 0-players, which refers to the distribution based on the role of 0-players. It means that removing a valueless player should not impact the bargaining power of other players, defined as[83],[84]:

$$\phi_i(N, v) = (N \backslash k, v) \text{ for any } i \in N \backslash k,$$
$$\text{s. t. }, k \in N \qquad (16)$$

The distribution of values can be summarized into two main concepts: the Shapley value and egalitarianism, or equal division. The Shapley value distributes based on individual value, while egalitarianism focuses on the value of the group, distributing an equal one-nth to everyone as part of forming a coalition. Most values are attempts to compromise between these two values, leading to the development of ENIC (egalitarian non-individual contribution) characteristics, which consider individual contributions before equitably distributing the surplus.

The upcoming CIS, ENSC, ENPAC, and ENBC values all originate from this characteristic and possess the axiom of Relatively Invariant under Strategic Equivalence (RISE). This axiom, usually satisfied along with ETP, indicates that a solution is standard and is defined as[82],[85]:

$$\phi(j) = \alpha \phi(k) + \beta,$$
$$\text{s. t., } \qquad \forall S \in N, \ j, k \in v(S),$$
$$\alpha > 0 \text{ and } \beta \in \mathbb{R} \qquad (17)$$

*B-2. Axioms of Values in Cooperative Game Theory*

To determine values in cooperative games, understanding the value of each coalition and the value of players within the coalition is essential, which requires a characteristic function. A commonly used method for this purpose is the Bankruptcy Problem. The Bankruptcy Problem is a method for dividing limited resources. In this context, each coalition takes the value after excluding the demands of players not in the coalition. This is mathematically represented as[86]:

$$V(C) = \max \left( 0, E - \sum_{i \in C} d_i \right) \qquad \forall C \subseteq N \qquad (18)$$

Here, $V(C)$ represents the maximum amount of resources that all players or players within coalition $C$ can take. The coalition $C$ is determined by the consensus of the players. $E$ denotes the quantity of limited resources, which must satisfy $E \leq \sum_{i \in N} d_i$, Typically, $(\sum_{i \in N} d_i) * 0.8$ is used as the value of $E$. $d_i$ represents the demand, and $N$ represents the set of all players. Once the value of each coalition $C$ is assessed, the value of each player's contribution within the coalition can be calculated.

**The Shapley Value** is recognized as one of the most rational methods for distribution in cooperative games. It was among the first to appear in cooperative games and is defined as the weighted average of the players' marginal contributions. The Shapley Value satisfies Efficiency (E), Linearity (L), Symmetry (S), and Null Player Out (NPO). As mentioned earlier, since it's determined by marginal contributions, it focuses on the capabilities of each member and is more concentrated on individual performance than the coalition as a whole. The marginal contribution, defined as , $v(C \cup \{i\}) - v(C)$, represents the contribution of a player in a coalition. The weight, defined as $\frac{|C|!(|N|-|C|-1)!}{|N|!}$, represents the number of cases for that coalition. The Shapley Value can be expressed with the following formula[87]:

$$\psi_i(v) =$$
$$\sum_{C \in (N - \{i\})} \frac{|C|!(|N|-|C|-1)!}{|N|!} \big( v(C \cup \{i\})$$
$$- v(C) \big) \qquad (19)$$

Alternatively, for all players $i \in N$, and all permutations $\pi \in \Pi$, let $M_{\pi,i} = \pi^{-1}(\{1, 2, \dots, \pi(i)\})$, then $\psi_i(v) = \sum_{\pi \in \Pi} \frac{v(M_{\pi,i}) - v(M_{\pi,i} - i)}{|N|!}$ is equivalent to formula (19). Egalitarianism can be represented as $ED_i(v) = \frac{v(N)}{|N|}$.

**The Banzhaf Value** is commonly used in voting games to analyze the influence of players based on their weights. Like the Shapley Value, it's defined by the vector of players' marginal contributions, but the weighting is distributed equally across all coalitions with a weight of $1/2^{\wedge}(|N|-1)$. The Banzhaf Value satisfies Linearity (L), Symmetry (S), Anonymity (AN), and NPO, and is expressed as[88],[89]:

$$B_i(v) = \sum_{C \in (N-\{i\})} \frac{1}{2^{|N|-1}} \big( v(C \cup \{i\}) - v(C) \big) \qquad (20)$$

**The Solidarity Value** aims for an equitable distribution among coalition members, contrasting with the Shapley Value which is based solely on individual value. It distributes an individual's contribution, $\Delta(v,C)$, equally among members of the same coalition and calculates the weight for the coalition similar to the Shapley Value. The Solidarity Value satisfies Efficiency (E), Linearity (L), Symmetry (S), and NPO[90]:

$$So_i(v) = \sum_{C \subseteq N : i \in C} \left( \frac{(|N| - |C|)!(|C| - 1)!}{|N|!} \Delta(v, C) \right),$$
$$\text{s.t.} , \Delta(v, C) = \frac{1}{|C|} \left( \sum_{i \in C} \big( v(C) - v(C - \{i\}) \big) \right) \qquad (21)$$

Also known as the equal surplus solution, the **CIS (Center of the Imputation Set) Value** assigns individual values to players and then equally distributes the surplus in the coalition among all players. It's crucial for egalitarianism and useful in games where either complete cooperation or failure of cooperation among all players is possible[84]. The CIS Value, originating from the ENIC characteristics, satisfies Efficiency (E), Symmetry (S), Additivity (AD), NPO, and RISE. It is expressed as[84]:

$$CIS_i(N, v) :=$$
$$v(\{i\}) + \left( \frac{1}{|N|} \left[ v(N) - \sum_{j \in N} v(\{j\}) \right] \right), \qquad (22)$$
$$\text{s.t.} , i \in N$$

Known as a dual value of CIS, the **ENSC (Egalitarian Non-Separable Contribution Value)** allocates value to all players based on their contribution in the grand coalition and then equally distributes the remaining surplus[91],[92]. The ENSC Value satisfies Efficiency (E), Symmetry (S), Additivity (AD), NPO, and RISE:

$$ENSC_i(v) :=$$
$$\varepsilon_i(N, v) + \left( \frac{1}{|N|} \left[ v(N) - \sum_{j \in N} \varepsilon_j(N, v) \right] \right), \qquad (23)$$
$$\text{s.t.} , \varepsilon_i(N, v) = \big( v(N) - v(N - \{i\}) \big)$$
$$and , i \in N$$

**The ENBC (Egalitarian Non-Banzhaf Contribution) Value**, similar to ENSC, assigns individual values to players and equally distributes the remaining surplus. The difference with ENSC is that the individual value is calculated using the Banzhaf Value[85],[92]:

$$ENBC_i(N, v) :=$$
$$\Gamma_i(N, v) + \left( \frac{1}{|N|} \left( v(N) - \sum_{j \in N} \Gamma_j(N, v) \right) \right),$$
$$\text{s.t.} , \Gamma_i(N, v) = \qquad (24)$$
$$\frac{1}{2^{|N|-1}} \left( \sum_{C \subseteq (N-\{i\})} \big( v(C \cup \{i\}) - c(C) \big) \right)$$

**The ENPAC (Egalitarian Non-Pairwise Averaged Contribution) Value** allocates individual value based on pairwise contributions in the grand coalition and then equally distributes the remaining surplus. It satisfies Symmetry (S), Additivity (AD), and RISE[92]:

$$ENPAC_i(v) :=$$
$$\Upsilon_i(N, v) + \left( \frac{1}{|N|} \left( v(N) - \sum_{j \in N} \Upsilon_j(N, v) \right) \right),$$
$$\text{s.t.} , \Upsilon_i(N, v) = \qquad (25)$$
$$v(N) - \left( \frac{1}{|N| - 2} \left( \sum_{j \in N-\{i\}} v(N - \{i, j\}) \right) \right)$$

The introduced ENIC values differ in how they distribute individual values of players but share the concept of equitably distributing the surplus.

**The Consensus Value** generalizes the standard solution of 2-player games to n-player cases, assigning individual values and satisfying Efficiency (E), Symmetry (S), and Additivity (AD). It aims to improve on the limitations of the Shapley Value and CIS Value by calculating marginal contributions through a generalized standard solution[93]. The standard solution for 2-player games, where players equally share the surplus created by their cooperation, is defined as follows for players $N=\{A,B\}$ and characteristic function $v$ as $v(\{A\}), v(\{B\}), v(\{A, B\})$[93]:

$$ST_A =$$
$$v(\{A\}) + \frac{v(\{A, B\}) - v(\{A\}) - v(\{B\})}{2} \text{ and}$$
$$ST_B = \qquad (26)$$
$$v(\{B\}) + \frac{v(\{A, B\}) - v(\{B\}) - v(\{A\})}{2}$$

Generalizing this to n-player games, the Consensus Value is calculated by applying the chain rule to all permutations of players until one player remains, defining the value as[93]:

$$CoV_i(v) =$$
$$\frac{1}{|N|}\left(\sum_{j\in(N-\{i\})} CoV\left((N-\{j\}),v^{-j}\right) + (v(\{i\})\right.$$
$$\left. + \frac{v(N)-v(N-\{i\})-v(\{i\})}{2}\right), \quad (27)$$

$$\text{s.t.,} \quad v^{-i}(C) =$$
$$\begin{cases} v(C) & \text{if } C \subsetneq N-\{i\} \\ v(N-\{i\}) + \dfrac{v(N)-v(N-\{i\})-v\{i\}}{2} & \text{if } C = N-\{i\} \end{cases}$$

## IV. PROPOSED METHOD

The objective of this paper is to find a method that distributes weights while maintaining and reflecting the characteristics of each classifier to the greatest extent possible. Relying on a single criterion for weight determination does not sufficiently reflect the diverse information obtainable from classifiers. In the proposed method, the confusion matrix, which is information about the basic classifiers obtainable at the ensemble stage, is assumed to represent the knowledge learned by the classifier about the data. The method suggests calculating the final weights while maintaining the ability to reflect the importance of the various information characteristics derived from this confusion matrix.

This approach involves a more comprehensive consideration of the classifier's performance, not limited to traditional metrics like accuracy or error rate but extending to a nuanced understanding of its behavior in different scenarios represented by the confusion matrix. The final weighting scheme, therefore, aims to account for a richer set of data-derived insights, potentially leading to a more robust and effective ensemble learning model.

### A. Multi-Criteria Decision Making Technique

In the proposed method, the VIKOR approach from the field of MCDM (Multi-Criteria Decision Making) is used to maintain the diverse characteristics of information and adjust their importance. VIKOR is a solution developed to address decision-making in situations with multiple conflicting considerations. It falls under the cooperative game strategy involving compromise and negotiation[66]. VIKOR is known for its high sensitivity among MCDM methods and is used in various fields such as risk management and water resource distribution[94]-[96]. The results from VIKOR typically represent a set of compromises, forming a basis for negotiation[66].

To implement the VIKOR method, the proposed approach utilizes performance evaluation metrics of classifiers. In the case of multi-class classifiers, performance varies across classes and also changes depending on the evaluation metric used. The proposed method considers these differences by focusing on the types of evaluation metrics and the disparity in class-wise evaluations.

The VIKOR method aims to select the point closest to the ideal solution. Utilizing the performance metrics of classifiers for this purpose is depicted in Figure 1. This approach ensures that the diverse and sometimes conflicting aspects of classifier performance are comprehensively considered, leading to a more balanced and effective decision-making process for weight allocation in ensemble models.

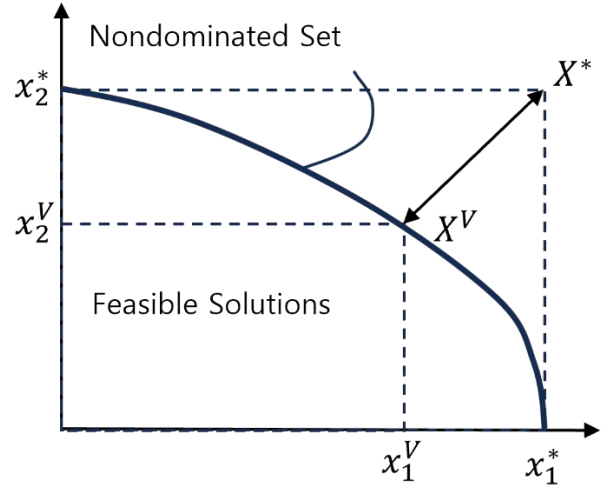

Figure 1: Classifier's VIKOR Solution.

In the figure, $x_i^j$ represents the performance achieved by classifier '$i$' on criterion '$j$', while $x_*^j$ and $x_-^j$ denote the best and worst performances, respectively, on that criterion. $X^*$ represents the ideal performance level (relative), and $X^V, x_j^V$ represent the solution selected through the VIKOR method. The method depicted in the non-dominated set of the figure involves finding the point closest to $X^*$. From the perspective of classifier ensembles, this means identifying the point where the performance achieved on various criteria by different classifiers is closest to the ideal value. The evaluation metrics used include ACCURACY, PPV (Positive Predictive Value), NPV (Negative Predictive Value), TPR (True Positive Rate), and TNR (True Negative Rate), utilizing individual assessment values for each class instead of average values or F-1 scores.

"Firstly, the performance for each class is quantified using the VIKOR method. The quantification of performance per evaluation metric is conducted first, followed by the quantification of relative evaluation for classifiers considering all evaluation metrics. Considering $m$ classes, $n$ classifiers, and $s$ performance indices, each classifier's performance in $C_1, \ldots, C_n$ is represented as $x_i^j(k)$ where $i = 1, \ldots, n, j = 1, \ldots, m, k = 1, \ldots, s$. This can be presented in $k$ matrices of $i \times j$, and for each $k$, formulating $i$ as alternatives and $j$ as criteria, the VIKOR method yields $k$ matrices of $i \times 1$. This value is represented as the index $y_i^k(k)$ for the kth performance of classifier $i$. This is expressed as follows, The VIKOR method for the kth iteration is conducted according to the following formula, based on formulas (8)-(10):

$$y_i^k = \left(Q_{max}(k) + Q_{min}(k)\right) - Q_i(k),$$

$$\text{s.t.,} \quad Q_i(k) = \frac{v\left(S_i(k) - S^*(k)\right)}{S^-(k) - S^*(k)}$$
$$+ (1-v)\frac{\left(R_i(k) - R^*(k)\right)}{\left(R^-(k) - R^*(k)\right)},$$

$$\text{and } S_i(k) = \sum_{j=1}^{m} \frac{p_j\left(x_*^j(k) - x_i^j(k)\right)}{x_*^j(k) - x_-^j(k)}, \tag{28}$$

$$\text{and } R_i(k) = \max_{j=1,\dots m}\left[\frac{p_j\left(x_*^j(k) - x_i^j(k)\right)}{x_*^j(k) - x_-^j(k)}\right].$$

In the kth evaluation of performance indices, the weights p_j for each criterion considered in the game are given inversely proportional to the number of classes to account for imbalanced classes. The weight for each criterion corresponding to a class is given by the following formula, based on the gamma distribution[97],[98]. In the proposal, the weights are simply given inversely proportional to the number of classes, but these weights can vary depending on the information to be considered."

The weight p_j for each criterion is calculated using the formula:

$$p_j = m \cdot \exp\left(-m \cdot w_j'\right),$$
$$\text{s.t.,} \quad w_j' = \frac{w_j}{\sum_{j=1}^{m} w_j} \tag{29}$$

Here, $w_j (j = 1, \dots, m)$ represents the number of instances for each class $j$.

Next, these $y_i^k$ values are re-formulated and computed using VIKOR to calculate the evaluation vector $z_i$, which serves as the basis for negotiation. The formula is as follows:

$$z_i = \left(\max_{i=1,\dots N}(Q_i) + \min_{i=1,\dots N}(Q_i)\right) - Q_i,$$

$$\text{s.t.,}$$
$$Q_i = \frac{v(S_i - S^*)}{S^- - S^*} + (1-v)\frac{(R_i - R^*)}{(R^- - R^*)},$$

$$\text{and } S_i = \sum_{j=1}^{m} \frac{p_j\left(y_*^k - y_i^k\right)}{y_*^k - y_-^k}, \tag{30}$$

$$\text{and } R_i = \max_{j=1,\dots m}\left[\frac{p_j\left(y_*^k - x_i^k\right)}{x_*^k - x_-^k}\right]$$

In this case, $p_j$ can be the same value or adjusted according to the user's assessment of importance. In the proposed method, it is assumed that all performance indices have equal importance, and the same value is assigned.

In the VIKOR method, as the calculation involves continuous scaling, the absolute value of the weight $p_j$ does not influence the outcome if there is no relative difference in values. Therefore, any arbitrary number can be used for this value.

While the proposed method uses only the performance indices of classifiers, it can evaluate classifiers based on other criteria if they use different features or if there are measures to assess their uniqueness. In such cases, the corresponding importance can be applied to assess the classifiers."

Once the evaluation vector is defined by formula (30), considering each class and evaluation metric, this vector is then used to calculate how each classifier should exert its influence. This is done using values derived from cooperative game theory. The VIKOR method focuses on ranking and selection, offering compromise solutions for conflicting criteria. For the problem of efficient distribution, solutions belonging to cooperative games are used, such as value analysis in weighted voting games and resource distribution[88],[99].

## B. Weight Distribution Through Value Calculation

Using the values obtained through formula (30) and the characteristic function, the calculation and normalization of values according to the methods of value calculation described later yield the weights r_i corresponding to the resources that should be allocated to each classifier. The characteristic function for calculating the value is given as a bankruptcy problem according to formula (18):

$$v(O) =$$
$$\max\left(0, W - \sum_{i \notin C} z_i\right) \quad \forall O \subseteq E, \tag{31}$$
$$\text{where } W = 0.8\left(\sum_{i=1}^{n} z_i\right)$$

Here, $v(O)$ represents the value that can be taken by coalition $O \in E$, which is a subset of the set of classifiers forming the ensemble $E = \{C_1, \dots, C_n\}$.

Once the value for each coalition $O$ is evaluated, the value can be calculated by measuring the contribution of each player in the coalition. The method of calculating the value using $v(O)$ given by formula (31) is provided by formulas (19)-(27) as follows:

The calculation of the Shapley value by $v(O)$ is given by formula (19):

$$V_i^s(v) =$$
$$\sum_{O \in (E - \{i\})} \frac{|O|!\,(|E| - |O| - 1)!}{|E|!}\left(v(O \cup \{i\}) - v(O)\right) \tag{32}$$

The calculation of the Banzhaf value by $v(O)$ is given by formula (20):

$$V_i^B(v) = \sum_{O \in (E - \{i\})} \frac{1}{2^{|E|-1}}\left(v(O \cup \{i\}) - v(O)\right) \tag{33}$$

The calculation of the Solidarity value by v(O) is given by formula (21):

$$V_i^{So}(v)$$
$$= \sum_{O \subseteq E: i \in O}\left(\frac{(|E| - |O|)!\,(|O| - 1)!}{|E|!}\Delta_i(v)\right),$$

$$\text{s.t.,} \Delta_i(v) = \frac{1}{|O|}\left(\sum_{i \in O}\left(v(O) - v(O - \{i\})\right)\right) \tag{34}$$

The calculation of the CIS value by v(O) is given by formula (22):

$$V_i^{CIS}(v) =$$

$$v(\{i\}) + \left(\frac{1}{|E|}\left[v(E) - \sum_{j \in E} v(\{j\})\right]\right), \qquad (35)$$

$$\text{s.t., } i \in E$$

The calculation of the ENSC value by v(O) is given by formula (23):

$$V_i^{ENSC}(v) =$$

$$\varepsilon_i(v) + \left(\frac{1}{|E|}\left[v(E) - \sum_{j \in E} \varepsilon_j(v)\right]\right), \qquad (36)$$

$$\text{s.t., } \varepsilon_i(v) = \left(v(E) - v(E - \{i\})\right) \text{ and } i \in E$$

The calculation of the ENPAC value by v(O) is given by formula (25):

$$V_i^{ENPAC}(v) :=$$

$$Y_i(v) + \left(\frac{1}{|E|}\left(v(E) - \sum_{j \in N} Y_j(v)\right)\right),$$

$$\text{s.t., } Y_i(v) = \qquad (37)$$

$$v(E) - \left(\frac{1}{|E| - 2}\left(\sum_{j \in (E - \{i\})} v(E - \{i, j\})\right)\right)$$

The calculation of the ENBC value by v(O) is given by formula (24):

$$V_i^{ENBC}(v) :=$$

$$\Gamma_i(v) + \left(\frac{1}{|E|}\left(v(E) - \sum_{j \in N} \Gamma_j(v)\right)\right),$$

$$\text{s.t., } \Gamma_i(v) = \qquad (38)$$

$$\frac{1}{2^{|E|-1}}\left(\sum_{O \subseteq (E - \{i\})}\left(v(O \cup \{i\}) - c(O)\right)\right)$$

The calculation of the Consensus value by v(O) is given by formula (27):

$$V_i^{Con}(v) =$$

$$\frac{1}{|E|}\left(\sum_{j \in (E - \{i\})} V^{Con}\left((E - \{j\}), v^{-j}\right) + (v(\{i\})\right.$$

$$\left. + \frac{v(N) - v(E - \{i\}) - v(\{i\})}{2}\right), \qquad (39)$$

$$\text{s.t., } v_{-i}(O) =$$

$$\begin{cases} v(O) & \text{if } O \subseteq E - \{i\} \\ v(E - \{i\}) + \dfrac{v(E) - v(E - \{i\}) - v(\{i\})}{2} & \text{if } O = E - \{i\} \end{cases}$$

In the case of the ENIC value, negative values can be assigned to the 0-player depending on the distribution of $z_i$[85]. Thus, when the value of classifier $i$ according to value $k$ is $V_i^k$, the final weight $r_i$ is defined by the following formula:

$$r_i^v = \frac{a_i^v}{\sum_{i=1}^n a_i^v} \qquad (40)$$

$$\text{whare } a_i^v = \max(0, V_i^v)$$

Once the final weights are calculated using the introduced values, these weights are used to conduct voting, and in the case of weighted voting, the soft outputs of classifiers are utilized. This is because the committee approach works well when based on probabilistic estimates of base classifiers[100]. Thus, in cooperative game-based voting, soft outputs are used, and the ensemble according to value k is defined as follows:

$$E_{cgv} = \underset{j=1,\dots,m}{\operatorname{argmax}} \sum_{i=1}^n r_i^v So_i^j,$$

$$\text{s.t., } r_i^v = \frac{\max(0, V_i^v)}{\sum_{i=1}^n \max(0, V_i^v)} \qquad (41)$$

In the final analysis stage, the performance differences according to the values are examined, and which values contribute more to performance enhancement in the cooperative game situation formed by classifiers is investigated. Analyzing the performance differences according to values leads to interpretations of how to distribute weights in the voting environment to make the right decisions.

If values that focus on individual contributions, such as the Shapley value and the Banzhaf value, show high performance, this means that the indicators considered in the ensemble have been sufficiently considered. Conversely, if values that focus on the value of cooperation itself show higher performance, it indicates that unconsidered factors exist and that the mere formation of coalitions leads to performance improvement.

The reason for considering several values as candidates stems from uncertainty. Cooperative games are considered as N-person decision-making games[101], and in these N-person decision problems, there are many uncertainties and unknowns, making predictions always imprecise due to differences in players' perceptions, inputs, and judgments[102]. To analyze this uncertainty, different values with slightly different concepts are applied, allowing for the comparison of performance according to values and the adoption of the value that shows the highest performance.

### C. Application of the Proposed Method

To summarize the proposed ensemble framework based on cooperative games, it involves evaluating pre-information obtained from training and testing of base classifiers through a cooperative game in a multi-criteria manner, distributing weights as resources, and then evaluating the performance of the ensemble using a test set. This process is illustrated in Figure 2.

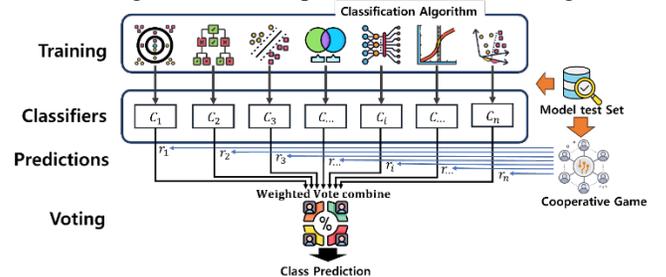

Figure 2: Classifier Ensemble Framework Based on Cooperative Games

The proposed method corresponds to the cooperative game part in the diagram, determining the weights of classifiers through a cooperative game using the information of each classifier. The cooperative game reflecting various features for the classifier weight $r_i$ is summarized in the following steps:

1. Selection and identification of features for the ensemble classifiers.

At this stage, the choice of classifiers to use and the features of the classifiers forming the ensemble are decided. The proposed method considers the class and performance indicators as features, noting that classifiers have various performance indicators and that, in the case of multi-class, performance indices differ for each class.

2. Simultaneous consideration of class-specific evaluations.

Here, classes are considered as criteria, and a global criterion for comprehensive evaluation is created using the MCDM method. The weight for each criterion is inversely proportional to the number of instances in that class, and these weights can be adjusted appropriately if there are considerations like error costs.

3. Simultaneous consideration of evaluations for each performance indicator.

Applying the process in step 2 to each performance indicator yields a value integrating class-specific evaluations as a result of the game for each performance indicator. This value is then subjected to the MCDM method with each performance indicator as criteria. The weight for each criterion is given equally, adjustable according to the environment. If other factors besides performance indicators are considered, they can be added with different weights.

The VIKOR method used in steps 1 and 2 represents values as a scaled distance from the ideal point, showing a distance of 0 for the best alternative in all aspects. Thus, this value is inverted to use as the evaluation value for classifiers.

4. Calculation of value.

The results from step 3 are multi-criteria evaluations considering both classes and performance indicators. These results are then used to calculate values for distributing weights.

5. Determination of the value to use.

Performance according to different values is assessed, and weights are distributed based on the value with the highest performance.

## D. Experiment

For the dataset, we used datasets from OPENML-CC18, which are frequently used for benchmarking purposes[103]. We compared performance using multi-class datasets available on OPENML and data with features extracted from CIFAR-10. Descriptions of each dataset can be referenced on OPENML.

The dataset is divided into three sets: training set for model training, model test set for classifier evaluation, and ensemble test set for comparing ensemble performance. Stratified sampling is used for splitting the dataset, assuming that the collected dataset and the actual data distribution are identical.

As base classifiers, we used K-Nearest Neighbor (KNN), Decision Tree (DT), Support Vector Machine (SVM), Naïve Bayes (NB), Artificial Neural Network (ANN/MLP), Quadratic Discriminant Analysis (QDA), and Logistic Regression (LR). The performance comparison metric used is accuracy (ACCURACY), which corresponds to the sum of the diagonal of the confusion matrix in multi-class classification.

The comparison involved the highest scores from existing weighting methods and the proposed method, and Table 1 compares the accuracy of the conventional voting method (soft voting; corresponding to formulas (3)-(7)) and the proposed weighted voting method.

Table 1: Accuracy Comparison by Method

| Data | Non-weight | SWV | RSWV | BWWV | QBWWV | WMV | Proposed Method |
|------|------------|------|------|------|-------|------|-----------------|
| Surface defects | 0.7422 | 0.7484 | 0.7319 | 0.7546 | 0.7402 | 0.7484 | **0.7634** |
| Mfeat Morphological | 0.71 | 0.726 | 0.728 | 0.722 | 0.722 | 0.724 | **0.732** |
| Mfeat Factor | 0.84 | 0.842 | 0.842 | 0.834 | **0.846** | 0.84 | **0.846** |
| Mfeat fourier | 0.836 | 0.838 | 0.838 | 0.844 | 0.832 | 0.838 | **0.85** |
| CIFAR-10 | 0.904 | 0.904 | 0.904 | 0.908 | 0.91 | 0.904 | **0.912** |

The experimental results observed that the proposed method showed higher accuracy improvement compared to other methods. This is presumed to be because the proposed method reflects more information that needs to be considered in the ensemble compared to other methods.

Table 2: Accuracy Comparison by Value

| Data | Shapley | Banzhaf | SO | CIS | ENSC | ENPAC | ENBC | CON |
|------|---------|---------|------|------|------|-------|------|------|
| Surface defects | 0.7572 | 0.7593 | 0.7572 | 0.7469 | **0.7634** | 0.7634 | 0.7613 | 0.7510 |
| Mfeat-Morphological | 0.728 | **0.732** | 0.73 | 0.728 | 0.726 | 0.726 | 0.726 | 0.73 |
| Mfeat-Factor | **0.846** | **0.846** | 0.838 | 0.844 | 0.844 | 0.844 | **0.846** | 0.842 |
| Mfeat-fourier | 0.848 | 0.848 | 0.842 | **0.85** | 0.844 | 0.844 | 0.846 | 0.838 |
| CIFAR-10 | 0.9063 | 0.9062 | 0.8985 | 0.8985 | **0.907** | 0.907 | 0.9069 | 0.9043 |

Table 2 compares the accuracy of the proposed method by value, where the difference in accuracy according to the value was not significant except for the Solidarity Consensus value. ENIC-Shapley and Banzhaf values showed higher performance and outperformed conventional methods. This suggests that in the ensemble, the distribution of weights without considering partial cooperation and focusing on individual contributions and overall fairness in the coalition is essential.

The experimental results showed that ENIC value, Banzhaf, and Shapley values demonstrated superior performance in sequence, indicating that in the distribution of weights in an ensemble, individual contribution and the fairness of the entire coalition play significant roles. The relatively lower performance of other values suggests that the value of partial cooperation is less related to ensemble performance, emphasizing the balance between contribution and fairness in the distribution of ensemble model weights.

The fact that the values showing the best performance vary across datasets indicates that not only the performance indicators considered in a multi-criteria approach but also several other factors in the distribution of ensemble weights influence the outcome. This implies that besides performance, differences in model characteristics, dataset structures, and various environmental factors can affect the performance of an ensemble.

## V. CONCLUSION

In this study, we proposed a novel approach to distribute weights in traditional weighted voting ensembles by viewing it from a multi-criteria perspective and considering various environments. We used the VIKOR method and value-based cooperative game theory for this purpose. Compared to traditional methods, this approach simultaneously considers multiple environments, optimizing weights to effectively overcome issues like class imbalance, overfitting/underfitting, and limitations of hypothesis space. The experimental results showed that our proposed method significantly outperformed the traditional methods, indicating that performance enhances when multiple pieces of information available from classifiers are considered simultaneously. An important aspect here is how the significance is assigned to each class and performance indicator during the consideration process, as this can greatly alter the extent of performance improvement.

The experimentation with various values for weight distribution revealed significant differences in weight distribution and performance. Specifically, ENIC, Banzhaf, and Shapley values demonstrated better performance, while Solidarity and CONSENSOUS values were less effective. This suggests that the initial criteria considered in the VIKOR method were well-reflected in the classifier diversity. The ENIC value, which assigns individual value in a specific way and evenly distributes the remainder, implies that the best performance under this value occurs in scenarios where the considered information is insufficient, and uncertainties due to incomplete information are equally weighted. Conversely, the low performance of CONSENSOUS and Solidarity values indicates that partial cooperation has little impact on performance, highlighting the importance of cooperation in a large coalition.

While only performance indicators were considered in the proposed method using the VIKOR method, future research could explore additional indicators and methods. Approaches like a game-theoretic perspective on the diversity and importance of classifiers, and correlation analysis offer alternative numerical methods. Applying these approaches to the proposed method, as well as considering other game-theoretic solutions like negotiation solutions in multi-criteria situations, could be worthwhile. This requires a deeper understanding of network decision theories like AHP, ANP, and a more sophisticated reflection on game theory. Additionally, research is needed on methods to calculate and incorporate the importance of information approached in various ways. Such comprehensive approaches are expected to maximize the performance of ensemble learning and explore applicability in diverse environments.